# Multi-Modal LLM based Image Captioning in ICT: Bridging the Gap Between General and Industry Domain

Lianying Chao[1,*], Kai Zhang[1,*], Haoran Cai[1], Sijie Wu[1], Xubin Li[1] (✉), Xin Chen[1]

[1]GTS, Huawei Technologies Co., Ltd.

[*]Lianying Chao and Kai Zhang contributed equally to this work.

✉ lixinbin@huawei.com

**Abstract.** In the information and communications technology (ICT) industry, training a domain-specific large language model (LLM) or constructing a retrieval-augmented generation system requires a substantial amount of high-value domain knowledge. However, the knowledge is not only hidden in the textual modality but also in the image modality. Traditional methods can parse text from domain documents but don't have image captioning ability. Multi-modal LLM (MLLM) can understand images, but they do not have sufficient domain knowledge. To address the above issues, this paper proposes a multi-stage progressive training strategy to train a Domain-specific Image Captioning Model (DICModel) in ICT, and constructs a standard evaluation system to validate the performance of DICModel. Specifically, this work first synthesizes about 7K image-text pairs by combining the Mermaid tool and LLMs, which are used for the first-stage supervised-fine-tuning (SFT) of DICModel. Then, ICT-domain experts manually annotate about 2K image-text pairs for the second-stage SFT of DICModel. Finally, experts and LLMs jointly synthesize about 1.5K visual question answering data for the instruction-based SFT. Experimental results indicate that our DICModel with only 7B parameters performs better than other state-of-the-art models with 32B parameters. Compared to the SOTA models with 7B and 32B parameters, our DICModel increases the BLEU metric by approximately 56.8% and 20.8%, respectively. On the objective questions constructed by ICT-domain experts, our DICModel outperforms Qwen2.5-VL 32B by 1% in terms of accuracy rate. In summary, this work can efficiently and accurately extract the logical text from images, which is expected to promote the development of multi-modal models in the ICT domain.

## 1 Introduction

With the explosive popularity of general large language models (LLMs) [1] in recent years, domain-specific LLMs and intelligent search engines have also seen corresponding development. It is well-known that both LLMs and retrieval-augmented generation (RAG) require vast amounts of high-value data, which originates from web pages, academic papers, or domain-specific documents. However, current document parsing methods, such as Miner-U [2], GOT [3], and Vary [4], can only parse textual content, resulting in the loss of high-value information from images. Professional documents in the ICT domain contain numerous high-value flowcharts and signaling diagrams. The absence of these visual knowledge may lead to insufficient learning for LLM training and incomplete information in the vector database. Therefore, the key to addressing these challenges lies in accurately converting visual information into clear and precise textual representations.

Previous research has developed numerous image captioning methods for converting visual information into semantic representations. Early approaches primarily relied on template-based methods. For example, Kulkarni *et al.* [6] developed a template-driven image captioning framework by detecting the relationship between objects, attributes, and prepositional relationships within images. Then, they employed conditional random fields for joint inference to generate natural language descriptions. Although the descriptions generated by this manner are standard and grammatically correct, they rely on pre-defined templates and lack diversity and flexibility. Alternatively, retrieval-based methods can also generate an image caption through image matching mechanisms. These methods first store numerous image-caption pairs into the database, and then compare the input image and the images in the database to find the similar image and the corresponding descriptions. The descriptions are organized reasonably to generate the final description of the input image. A representative work developed by Rata *et al.* [8] integrated retrieved similar images and their descriptions rather than relying solely on the input image itself. In this architecture, the encoder employs pre-trained BERT [7] to jointly process the input image and retrieved textual descriptions, while the decoder synthesizes these multi-modal representations to generate final captions. However, the efficacy of this approach is limited by three critical factors: the scale of annotated datasets, the performance of retrieval algorithms, and the accuracy of similarity computation. Like template-based methods, retrieval-based methods also inherently suffer from limited diversity in caption generation.

The development of deep learning has given rise to learning-based image captioning methods. These methods typically employ an encoder-decoder architecture that integrates computer vision and natural language processing tasks. Vinyals *et al.* [8] first introduced deep learning techniques into the image captioning task via a CNN-LSTM framework, where a convolutional neural network (CNN) encodes visual features and long short-term memory (LSTM) decodes textual descriptions. Xu *et al.* [9] further introduced attention mechanisms into the decoder, including additive attention, stochastic hard attention, and deterministic soft attention. With the evolution of computational resources, pre-trained model-based captioning methods have emerged as a dominant paradigm. For example, Zhou *et al.* [10] developed a VLP framework by utilizing a unified multi-layer Transformer for joint encoding-decoding, enabling bidirectional image-text generation and comprehension. Li *et al.* [11] analyzed the problems of insufficient model structure and data sources in existing models, and made two contributions: (1) proposing a multi-mode encoder-decoder hybrid structure, which can effectively pretrain and transfer learning for multi-task; (2) proposing a bootstrapping method to enhance the training dataset. Experimental results indicate that these methods can effectively improve the diversity and accuracy of image description.

With the expansion of computing resources, multi-modal LLMs (MLLMs) with 10 billion and 100 billion parameters have emerged. GPT-4 and its upgraded GPT-4.1 [12] series have demonstrated wide-ranging application potential in many fields with their powerful reasoning and long text processing capabilities. Unfortunately, GPT-4 series are closed source and can only provide APIs externally. Domestic MLLMs are developing rapidly, and most of them are open source. Typical MLLMs mainly consist of three components: vision encoder, connector, and LLM-based decoder. The main parameters come from the LLM-based decoder. For example, Shanghai Artificial Intelligence Laboratory has developed an open-source MLLM named Intern-VL [13-14], which aims to achieve efficient and flexible multi-modal interaction by merging visual and textual information. Intern-VL series introduce a dynamic high-resolution strategy to solve the problem of excessive memory occupation caused by large images. Specifically, the input image is divided into multiple slices with 448×448 pixels to avoid the distortion and detail loss caused by forced resizing, which can support 4K resolution. Intern-VL enables a variety of high-level visual processing tasks such as visual question answering (VQA), document and chart parsing, and image and video analysis. Alibaba has released Qwen-VL series [15], which can process images, text, and bounding boxes as inputs, and output text and bounding boxes. The training of Qwen-VL series includes three stages, *i.e.*, pre-training, multi-task training, and instruction fine-tuning, to gradually improve its understanding and generation capabilities for specific tasks. In addition, there are many large MLLMs, such as LLaVA [16], Gemini [17], and BLIP-2 [18], and so on. However, the training data of these MLLMs mainly come from general data and do not have professional background knowledge in the ICT domain. As a result, the image understanding precision is limited in directly applying to the ICT field. For example, in the ICT domain, the results of MLLMs' understanding often include the wrong logic judgment in flowcharts, the unclear information identification, and the opposite information flow direction in signal diagrams, and so on.

To solve the above issues, this paper develops a Domain-specific Image Captioning Model (DICModel) with a three-stage progressive training strategy to enable the image parsing and understanding tasks in the ICT domain. The performance of DICModel is evaluated on the ICT-domain testing data based on the experts' experience. Specifically, in the first stage, the Mermaid tool synthesizes about 7K image-text pairs that are used for the supervised-fine-tuning (SFT) of DICModel. In the second stage, ICT-domain experts manually annotate about 2K high-quality image-text pairs to train DICModel. In the third stage, domain experts produce about 1.5K pairs of questions and answers based on the domain images to perform instruction-based SFT. The training image types maintain alignment with the critical image types in the ICT domain, primarily encompassing flowcharts and signal diagrams. DICModel employs the Qwen2.5-VL 7B as the base model. The training pairs of the first two stages include images and the corresponding parsing results, which aim to improve the parsing precision. The instruction-based SFT is to make DICModel follow the user's intention and to complete the correct understanding of the input image according to the instruction.

To better evaluate DICModel, we invite ICT-domain experts to manually annotate and question the domain images. First, experts annotate 100 critical images and the corresponding text via fixed templates to evaluate the image parsing capability. Then, the experts construct 600 objective questions based on the images, including 200 single-choice questions and 100 multiple-choice questions, to evaluate the image understanding capability. Experimental results indicate that, with the same number of parameters, the proposed method achieves better performance than other state-of-the-art methods: the BLEU metric is improved by 56.8%, and the accuracy of objective questions is improved by 10.3% compared to the second-best method. Even compared with larger-sized models such as Qwen2.5-VL 32B, DICModel has achieved 20.8% and 1% improvements in terms of the BLEU metric and accuracy rate of objective questions, respectively.

The main contributions of this work are summarized as follows:

- This work develops a multi-modal model with a progressive training strategy to parse and understand high-value images in the ICT domain. The training data of the multi-modal model comes from synthetic data and high-quality manually-annotated data, which can accurately parse images and answer complex questions according to the images.
- This work uses the Mermaid tool to synthesize large amounts of flowcharts and signal diagrams and combines an LLM-based method to generate expert-approved text styles, which can synthesize high-quality image-text pairs.
- This work constructs a comprehensive evaluation system using experts' experience, including image parsing results and VQA capabilities. On the evaluation set designed by experts, the proposed method achieves the highest performance in both parsing and understanding images.

## 2 Related Work

The image captioning task belongs to an interdisciplinary field that combines computer vision and natural language processing, and aims to enable computer systems to comprehend visual information and automatically generate textual descriptions. Due to the limited diversity of descriptions produced by template-based and retrieval-based methods, this paper primarily focuses on deep learning-based approaches for image captioning generation. As illustrated in Fig. 1, learning-based methods typically consist of two core components: a visual encoder that extracts image features and a language decoder that generates textual descriptions. The following content is to introduce the related work on the learning-based methods for image captioning.

### 2.1 Graph-based Methods

Graph structures can effectively correlate visual and textual information, and researchers have attempted to integrate graphs into image captioning tasks. For example, Yang *et al.* [20] proposed a scene graph auto-encoder (SGAE) that incorporates linguistic inductive biases into the encoder-decoder framework to generate more human-like descriptions. They validated the effectiveness of SGAE on the challenging MS COCO image captioning benchmark, achieving state-of-the-art performance. To address the difficulty of acquiring paired image-caption data, Gu *et al.* [21] developed a scene graph-based method for unpaired image captioning. To better align key visual and textual information, researchers have combined graph structures with attention mechanism to enhance the accuracy and diversity of generated captions. For example, Yao *et al.* [21] introduced a Graph CNN and LSTM architecture that integrates both semantic and spatial object relationships into the image encoder. While graph-based methods for image captioning can effectively capture objects and their relationships within images, they demand large amounts of high-quality annotated data containing detailed graph structures and relationships. Therefore, graph-based methods exhibit limited generalization capabilities and are difficult to adapt to new scenarios that significantly differ from the training data.

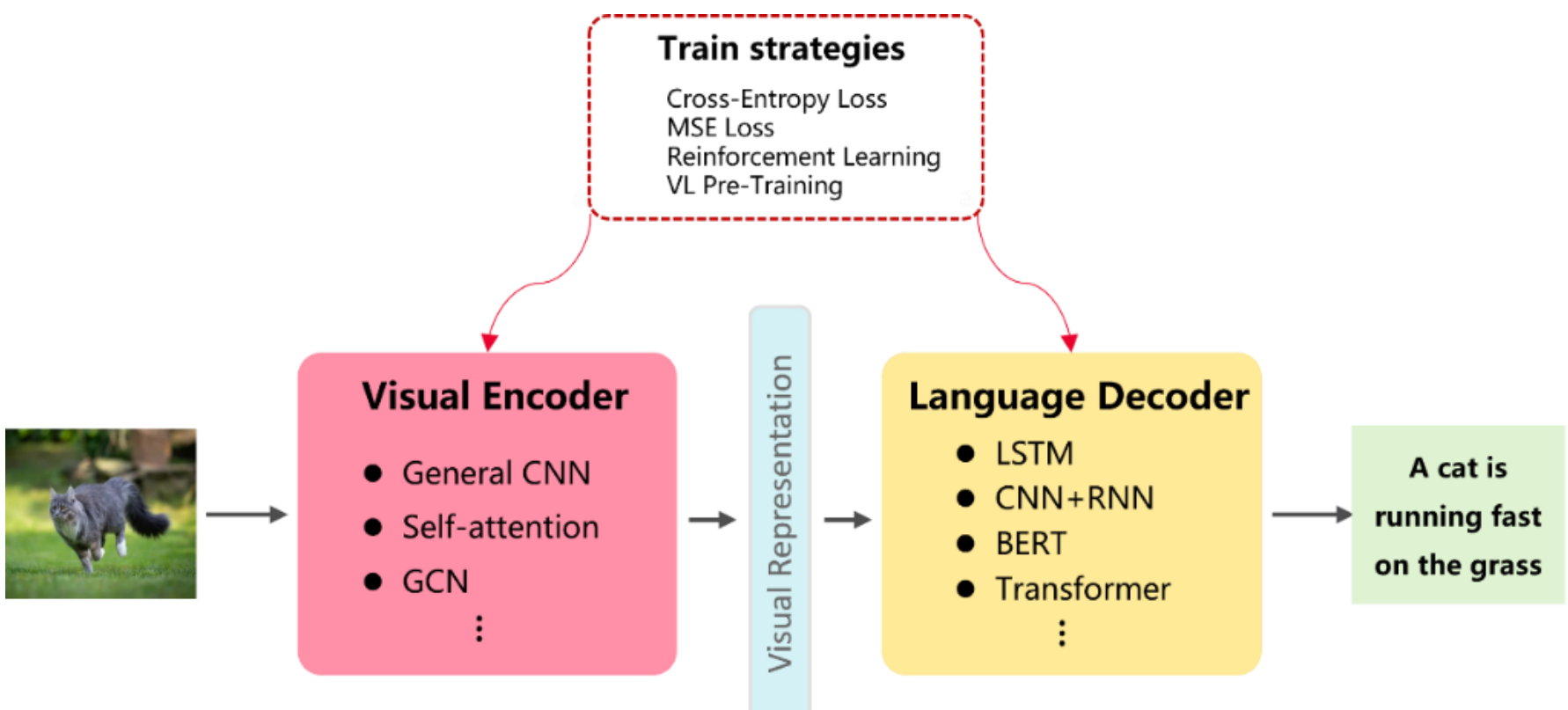

**Fig. 1.** Basic Paradigm of Learning-based Image captioning Methods.

### 2.2 Transformer-based methods

With the advancement of Transformer, researchers have discovered that the inherent attention mechanism of the Transformer makes it particularly suitable for text-related tasks. Consequently, many Transformer-based methods have emerged for solving image captioning tasks. For example, Herdade *et al.* [22] introduced an object-relation Transformer, which enhances spatial relationship modeling by explicitly incorporating geometric attention between detected objects. Huang *et al.* [23] developed an attention-on-attention module and applied it to both the encoder and decoder of their image-captioning model. Additionally, some methods integrate the Transformer with graphs to better leverage spatial relationships among image regions [24]. However, descriptions generated by Transformer may lack deep comprehension of spatial structures and object relationships within images, potentially resulting in text that is insufficiently accurate or lacks details.

### 2.3 MLLM-based methods

MLLMs are built upon the rapid development of LLMs and large-scale vision models, such as ViT, CLIP, and BLIP. These MLLMs typically consist of three components, *i.e.*, a visual encoder, a connector, and an LLM-based decoder [25]. Compared to training vision encoders from scratch, a common approach is to use models pre-trained on massive image-text pairs. For example, the visual encoder in CLIP can effectively convert image information into vector representations [26]. In recent years, MLLMs have experienced explosive growth. For example, GPT-4 supports multi-modal inputs and generates high-quality descriptions, which enables applications in image captioning, visual question answering, and multi-modal dialogue. Additionally, Google and DeepMind have developed Gemini [17] and Flamingo [27], respectively, both demonstrating strong image comprehension capabilities. Domestic MLLMs also exhibit strong competitiveness. Qwen-VL series, developed by Alibaba, supports image and text inputs, and can generate outputs in formats with text, images, and detection bounding boxes. Qwen-VL series performs exceptionally well on the OpenCompass [28] multi-modal benchmark, achieving comparable performance to GPT-4V and Gemini 2.0. Shanghai AI Laboratory developed Intern-VL series, which incorporates dynamic high-resolution processing strategies and supports 4K resolution. Additionally, Intern-VL can handle generation tasks in 110 languages. Other notable domestic models include the Hunyuan-Large Model [29] developed by Tencent, GLM-4 [30] developed by Zhipu AI, and Doubao [31] by Bytedance. Despite their strong performance in understanding natural images, these MLLMs exhibit limited capabilities in parsing ICT-domain images, *i.e.*, flowcharts and signal diagrams, because these diagrams are not included in their training data.

## 3 Methodology

### 3.1 Overall Scheme

This section introduces the overall framework for training DICModel. The training process primarily consists of three stages. The first stage is pre-SFT, which uses large-scale synthetic data to enable the model to learn the standard paradigms of image parsing. The second stage is post-SFT, which uses expert-annotated high-quality data to further acquire domain-specific knowledge in the ICT domain. The final stage is instruction-SFT, which utilizes visual question-answer (VQA) pairs constructed by experts to ensure comprehensive understanding of images and enhance compliance with user instructions. The following sections will provide a more detailed explanation of each training stage and the techniques employed.

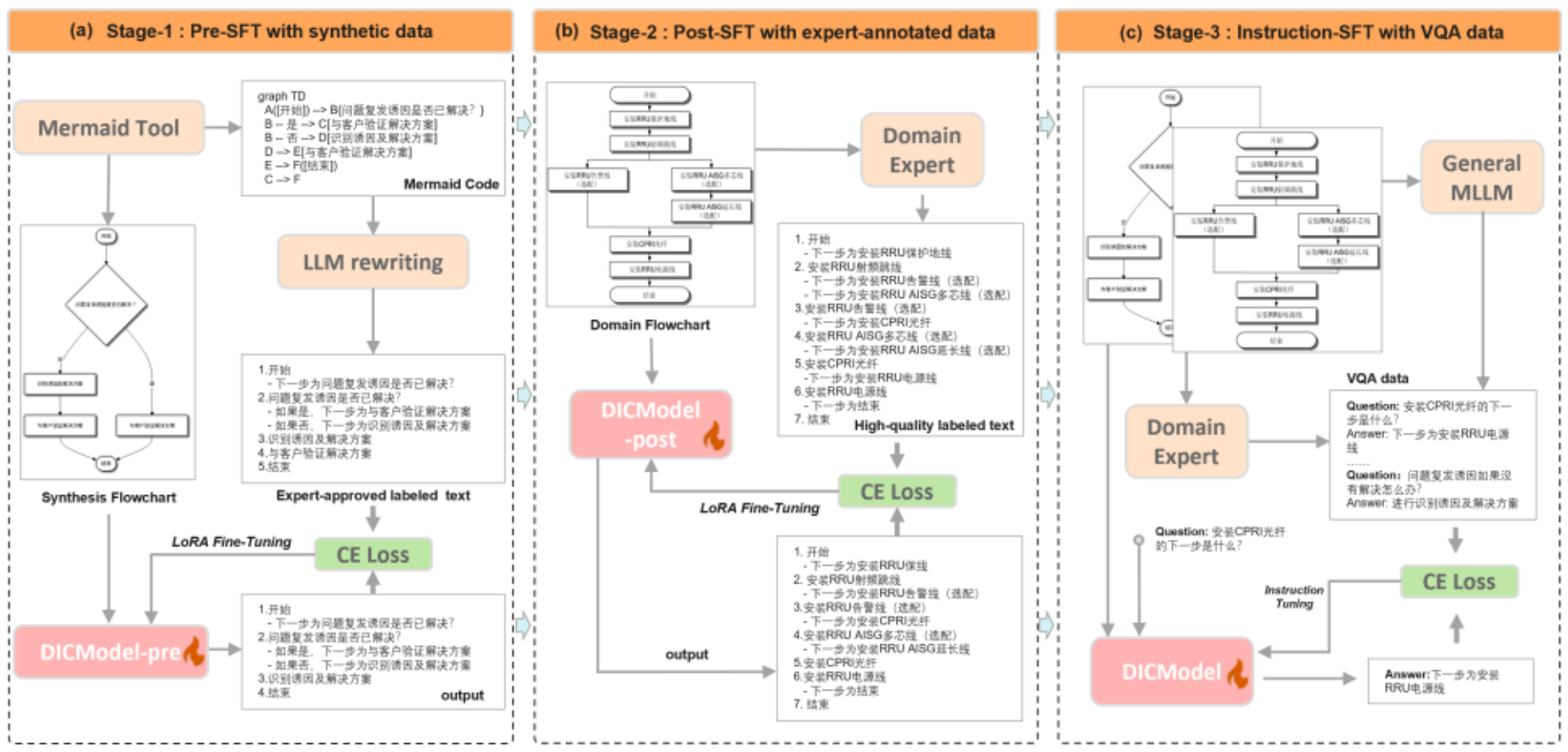

Fig.2. Progressive-training strategy for our DICModel.

### 3.2 Pre-SFT with Synthetic Data

As shown in Fig. 2(a), the first stage employs large-scale synthetic data to train DICModel, enabling it to learn structural information and a standardized text output format. To ensure efficient inference speed, we employ Qwen2.5-VL 7B as the base model. After the first-stage training, DICModel-pre is obtained and serves as the initial model for the second training stage.

Synthetic data is automatically constructed based on the Mermaid tool, which is an open-source diagram generation tool. The Mermaid toolbox utilizes markdown-like code syntax to create and modify various diagrams, *i.e.*, flowcharts, sequence diagrams, Gantt charts, class diagrams, state diagrams, pie charts, architecture diagrams, *etc*. As a text-driven diagramming method, Mermaid allows users to define diagram structures through textual coding, eliminating the need for manual adjustment of graphical elements. Moreover, the Mermaid toolbox can operate in diverse JavaScript-compatible environments, such as web browsers and markdown editors. The generation process of synthetic data can be formulated as:

$$I_s = f_{mermaid}(T_{sc}) \tag{1}$$

where $T_{sc}$ represents the Mermaid code, $f_{mermaid}(\cdot)$ represents the process of image synthesis using Mermaid code $T_{sc}$. $I_s$ is the synthesized image by using the structured Mermaid code $T_{sc}$.

However, $T_{sc}$ significantly deviates from standard annotation formats and cannot be directly utilized for training DICModel. To address the problem, our method employs Grok-3 [32] to convert $T_{sc}$ into standardized annotation text $T_s$. The output format of $T_s$ is optimized by the prompt engineering. The specific process can be formulated as:

$$T_s = P \circ f_{LLM}(T_{sc}) \tag{2}$$

where $T_S$ is the expert-approved annotation text, $P$ is the designed PROMPT, $f_{LLM}(.)$ represents the process of rewriting the Mermaid code by Grok-3 [32]. After generating the image-text pair $(I_s, T_s)$, DICModel model can be trained using the cross-entropy (CE) loss function that can be calculated as follows:

$$\mathcal{L}_{CE-pre} = -\sum_i T_{s,i} \log\,(T'_{s,i}) \tag{3}$$

where $T'_S = f_{DICModel-pre}(I_s)$, $T_{s,i}$ is the one-hot coding of $T_s$, $T'_S$ is the probability distribution predicted by DICModel-pre.

### 3.3 Post-SFT with Expert-Annotated Data

As shown in Fig. 2(b), after the first training stage, we further invite experts in the ICT domain to manually annotate high-quality image-text pairs from professional domain-specific documents. The main objective of the second stage is to enable DICModel to focus on the ICT domain. Leveraging their industry experience, experts design fixed text-annotation templates for each image category to help DICModel parse and comprehend images from an expert perspective. The expert annotation process can be formulated as:

$$T_R = f_{expert}(I_R) \tag{4}$$

where $I_R$ represents the domain image, $T_R$ is the experts-annotated text. Based on these annotated image-text pairs $(I_R, T_R)$, DICModel can be trained in the second stage. The calculation of loss function can be formulated as:

$$\mathcal{L}_{CE-post} = -\sum_i T_{R,i} \log(T'_{R,i}) \quad (5)$$

where $T'_R = f_{DICModel-post}(I_R)$, $T_{R,i}$ is the one-hot coding of $T_R$. The number of training epochs and learning rate in this stage are different from those in the first stage, which is detailed in the section 4.2.

### 3.4 Instruction-SFT with VQA Data

The first two training stages aim to enhance the image parsing capability of DICModel, which involves extracting textual content from images and semantically reconstructing the content according to expert-defined specifications. However, mere image parsing tasks remain relatively trivial for MLLMs with billions of parameters. We aspire for DICModel not only to parse images accurately but also to understand fine-grained details, local components, and global layouts.

To achieve this, as illustrated in Fig. 2(c), we leverage large-scale multi-modal models and domain experts to generate extensive VQA data to train DICModel in the third stage. This strategy enhances the model's instruction-compliance capability and image-understanding ability. The workflow can be formally represented as follows:

$$(Q_m, A_m) = f_{MLLM}(I_R) \quad (6)$$

$$(Q_r, A_r) = f_{Expert}(I_R) \quad (7)$$

where Q and A are the question and answer, $f_{MLLM}$ and $f_{Expert}$ are the process of question generation by MLLM and experts, respectively. $I_R$ is the ICT-domain image.

Therefore, the generated VQA data $(Q, A, I)$ can be used for training DICModel. The image $I$ and question $Q$ are used as the input of the model, and the answer $A$ is used as the label. The loss function still uses the CE loss function:

$$\mathcal{L}_{CE-instruct} = -\sum_i A_i \log(A'_i) \quad (8)$$

where $A'_i = [f_{DICModel}(Q, I)]_i$, $A_i$ is the one-hot coding of $A$. $\mathcal{L}_{CE-instruct}$ represents the loss function in the instruction SFT stage. Due to the limitations of computing resources, we use the LoRA technique to freeze some parameters of the model to achieve efficient fine-tuning during the three training processes.

## 4 Experiments

### 4.1 Data

Our method employs Qwen2.5-VL 7B as the base model. Since the training data of Qwen2.5-VL already include most publicly available datasets, we do not use open-source data to further train it, but instead employ automated tools and experts to annotate domain-specific images shown in Table 1, including flowcharts and signal diagrams.

**Table 1.** Training data of DICModel in three training stages.

| Training Stage | Pre-SFT | Post-SFT | Instruction-SFT |
|---|---|---|---|
| Data Type | Synthetic pairs | Real pairs | Real VQA |
| Tool | Mermaid+Grok-3 | Experts | Gemini2.0+ Experts |
| Number | 7233 | 2274 | 1573 |

| Domain | ICT style | ICT | ICT |
|---|---|---|---|

**Synthetic Data.** In the ICT domain, professional documents frequently utilize flowcharts and signal diagrams to guide first-line engineers. Consequently, our method synthesizes these two key image types using the Mermaid toolbox. Specifically, the Mermaid toolbox generates 7,233 images (3,200 flowcharts and 4,033 signal diagrams) and their corresponding Mermaid codes. We then leverage Grok-3 to convert Mermaid code into expert-approved annotation texts. Thus, the synthetic training dataset comprises 3,200 flowchart-text pairs and 4,033 signal diagram-text pairs.

**Domain Data.** To further enhance the model's performance in understanding ICT-domain images, domain experts manually annotated real-world images from the ICT domain. Experts annotate 2,274 image-text pairs, including 1,100 flowcharts and 1,174 signal diagrams.

**VQA Data.** VQA data originates from two sources: (1) MLLM generation, where Gemini 2.0 produces 1,000 Q-A pairs based on image content; (2) experts set the questions, where domain experts manually constructed 573 Q-A pairs leveraging their experience.

### 4.2 Training Details

We implement our network using Pytorch ver. 2.6.0, Python ver. 3.10.0, CUDA ver. 11.8, and Transformer ver. 4.49.0. Besides, other dependent packages are installed compatible with the above three. The computer configuration includes an Intel(R) Xeon(R) Gold 6151 CPU of 3.00 GHz, and eight NVIDIA GPUs Tesla V100 of 32 GB memory.

In the training process of DICModel, the batch size is set to 16, and gradient accumulation step is set to 4. The Adam optimizer is employed to update the DICModel parameters The model's precision is set to fp16. The learning rates for the first, second, and third stages are set to 1e-4, 1e-5, and 1e-4, respectively, and the number of training epochs is set to 2, 1, and 1.

### 4.3 Comparison methods

In this paper, we compare several SOTA multi-modal models, including Gemini-2.0 [17], Claude-3.5 [33], Phi-3.5-vision 4B [34], LLaVA-1.5 7B [35], Intern2.5-VL 7B [36], Qwen2.5-VL 7B [15], Intern2.5-VL 32B [36], and Qwen2.5-VL 32B [15]. Among them, Gemini-2.0 and Claude-3.5 are closed-source models, so they are tested via API. The remaining models are open-source, so we download the original weights and deployed and tested them on servers. To make the model outputs adhere as closely as possible to standard text formats, we optimized the prompt during the image parsing stage to specify the text output style.

### 4.4 Evaluation Metrics

To evaluate the image understanding ability from global to local contents, ICT-domain experts develop 200 single-choice questions and 100 multiple-choice questions based on their domain experience. The models' comprehension capabilities are ultimately evaluated by their accuracy in answering these questions. Eq. (8) and (9) represent the accuracy rates of the single-choice and multiple-choice questions, respectively. Eq. (10) is the average accuracy rates of all objective questions.

$$Prec_s = \frac{A_{s,r}}{A_{s,t}} \times 100\% \tag{8}$$

$$Prec_m = \frac{A_{m,r}}{A_{m,t}} \times 100\% \tag{9}$$

$$Prec_a = \frac{A_{s,r} + A_{m,r}}{A_{s,t} + A_{m,t}} \times 100\% \tag{10}$$

where $A_{s,r}$ and $A_{m,r}$ represent the correct answers, and $A_{s,t}$ and $A_{m,t}$ represent the total number of single-choice and multiple-choice questions, respectively. In this work, $A_{s,t}$ and $A_{m,t}$ is 200 and 100, respectively.

In addition to objective questions, this paper employs common quantitative metrics to better evaluate the parsing accuracy of comparison methods, including Consensus-based Image Description Evaluation (CIDEr), Metric for Evaluation of Translation with Explicit Ordering (METEOR), and Bilingual Evaluation Understudy (BLEU). These metrics are used for evaluating the text similarity between the output and the label.

# 5 Experimental Results

## 5.1 Results of Visual Question Answering

To validate the image understanding capabilities of comparison methods, this work invites the ICT-domain experts to design objective questions from multiple perspectives for domain-critical images. The experts produce a total of 300 questions, including 200 single-choice and 100 multiple-choice items. We evaluate the image understanding performance using the answer accuracy rates $Prec_s$, $Prec_m$ and $Prec_a$ formulated in *Eqs.* (8)-(10). Table 2 presents the accuracy rates of different methods.

Experimental results demonstrate that DICModel achieves the higher accuracy rate on single-choice questions compared to other SOTA models. For multiple-choice questions, DICModel achieves the second-highest accuracy rate, only lower than Intern2.5-VL with 32B parameters. In terms of average accuracy $Prec_a$, our DICModel outperforms the closed-source models (Claude-3.5 and Gemini-2.0) and the models with 32B parameter sizes (Qwen2.5-VL 32B and Intern2.5-VL 32B). With the same 7B number of parameters, our DICModel improves the average accuracy rate by 10.3% compared to the second-best Qwen2.5-VL model.

The superior performance of DICModel stems from the proposed three-stage training process, which progressively enhances image understanding from fine details to global content. In particular, the third training stage enables the model to achieve deeper comprehension of images, thereby significantly enhancs the answer accuracy. In summary, our method achieves SOTA performance in image understanding while maintaining a relatively small parameter size.

**Table 2.** Answer accuracy of comparison methods on expert-constructed objective questions. A total of 300 questions are used, including 200 single-choice and 100 multiple-choice items. Numbers in red and blue represent the best and second-best scores, respectively.

| Model Name | Size | $Prec_s$ | $Prec_m$ | $Prec_a$ |
|---|---|---|---|---|
| Claude-3.5 | API | 67.0% | 37.0% | 57.0% |
| | | *Multi-modal LLM based Image Captioning in ICT* | | |
| Gemini 2.0 | | 77.5% | 38.0% | 64.3% |
| Phi-3.5-vision | 4B | 48.5% | 10.0% | 35.7% |
| LLAVA-1.5 | 7B | 45.0% | 21.0% | 37.0% |
| InternLM-XC-2.0 | | 71.0% | 39.0% | 60.3% |
| Qwen2.5-VL | | 73.0% | 40.0% | 62.0% |
| **DICModel (ours)** | | **82.5%** | **52.0%** | **72.3%** |
| Intern2.5-VL | 32B | 79.5% | **55.0%** | **71.3%** |
| Qwen2.5-VL | | **80.0%** | 47.0% | 69.0% |

To further present the ability of image understanding, we select two flowcharts and two signaling diagrams from the ICT-domain documents. Fig. 3 presents the VQA results for the selected figures, including single-choice and multiple-choice questions.

As shown by Q1 in Fig. 3, only our DICModel and the closed-source Geimini-2.0 provide the correct answers, but other models fail to properly understand the image. This indicates that the two models can well understand the judgment logic of flowcharts and follow correct steps based on user's instructions. For the multiple-choice question Q2 about flowcharts, both our method and Intern2.5-VL-32B provide correct answers. As shown in Q3 and Q4, our method accurately interprets signal diagrams and identifies signaling information at specific nodes. For example, in Q4 when querying the number of signaling messages at node UE, our method returns the correct answer. Overall, our method achieves comparable even superior results to 32B models (Qwen2.5-VL and Intern2.5-VL) and closed-source models (Gemini 2.0).

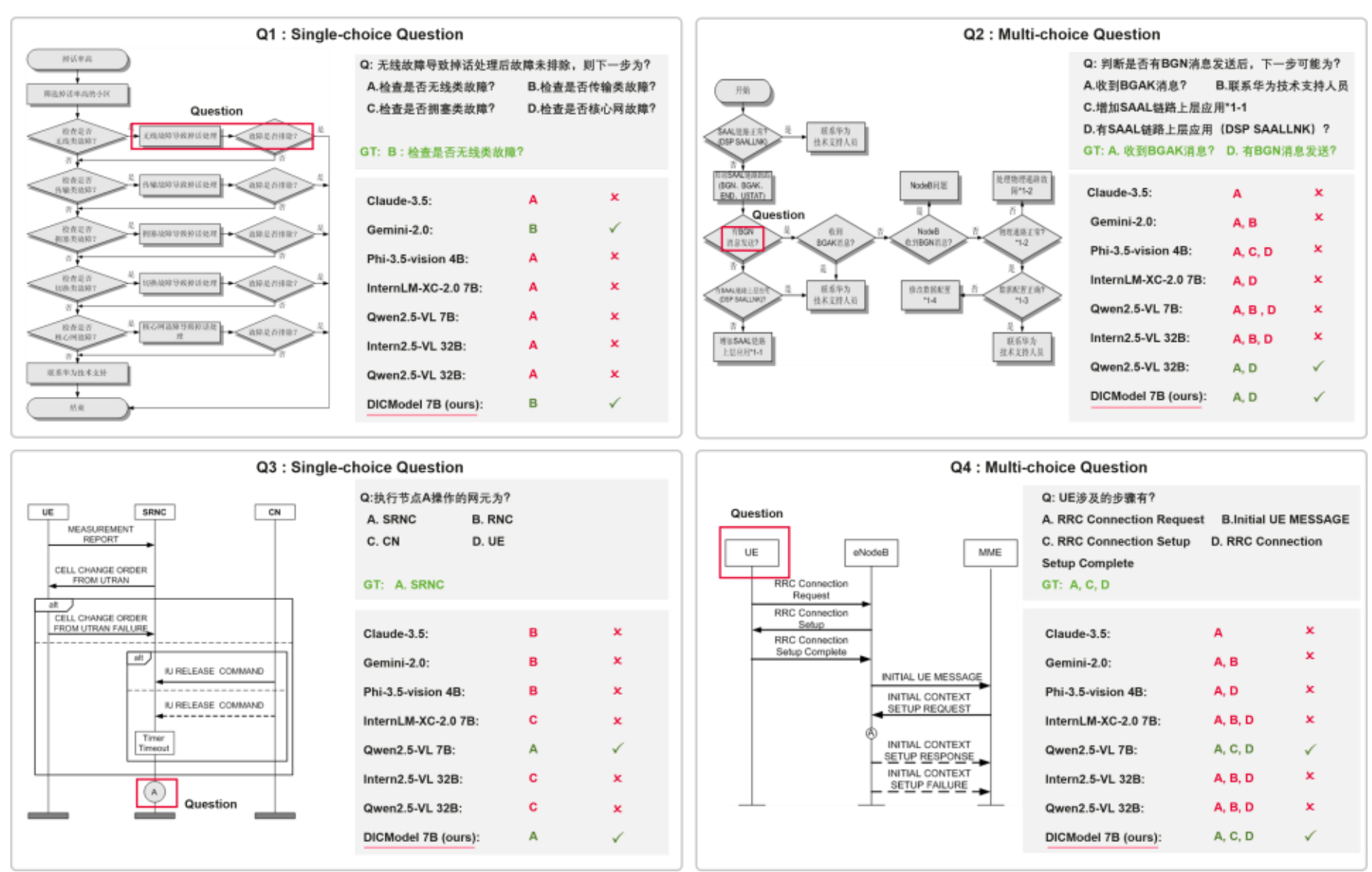

**Fig. 3.** Four critical images from ICT-domain documents, including two flowcharts (Q1: single-choice question; Q2: multiple-choice question) and two signal diagrams (Q3: single-choice question; Q4: multiple-choice question).

## 5.2 Results of Visual Parsing

This section evaluates the image parsing ability of comparison methods. Table 3 lists the quality metrics for 100 ICT-domain images, including 50 flowcharts and 50 signal diagrams. The textual content of these images is manually annotated by ICT-domain experts (as ground truth); model outputs are compared against the ground truth to compute quantitative metrics. Parsing quality is assessed using three common evaluation metrics (CIDEr, METEOR and BLEU) that comprehensively to measure the similarity between generated and reference text, diversity of generated text, and semantic accuracy of generated text.

As shown in Table 3, the experimental results demonstrate that the proposed DICModel achieves significantly better quantitative metrics compared to other comparison methods with the same parameter sizes (*e.g.*, LLaVA-1.5-7B, Intern2.5-VL-7B, and Qwen2.5-VL-7B). With equivalent 7B parameters, our method improves the BLEU score by 0.21 compared to the second-best model Qwen2.5-VL-7B. For the closed-source models (Claude-3.5 and Gemini-2.0), comparisons are performed by using their official APIs. Our DICModel achieves higher parsing accuracy than the two closed-source SOTAs. Furthermore, compared to the larger-scale models like Intern2.5-VL-32B and Qwen2.5-VL-32B, DICModel achieves higher CIDEr and BLEU scores while using only 22% of their parameters. In summary, the proposed method achieves the highest parsing accuracy among all compared methods. Moreover, our DICModel has a relatively small number of parameters, which can guarantee the image parsing efficiency.

**Table 3.** Quantitative assessments for the parsing precision of DICModel. 100 images from ICT domain are evaluated. Numbers in red and blue represent the best and second-best scores, respectively.

| Model Name | Size | CIDEr↑ | METEOR↑ | BLUE↑ |
|---|---|---|---|---|
| Claude-3.5 | API | 0.65 | **0.19** | **0.52** |
| Gemini-2.0 | | 0.66 | **0.19** | 0.50 |
| Phi-3.5-vision | 4B | 0.23 | 0.17 | 0.04 |
| LLAVA-1.5 | 7B | 0.26 | 0.12 | 0.26 |
| InternLM-XC-2.0 | | 0.50 | 0.17 | 0.35 |
| Qwen2.5-VL | | 0.53 | 0.16 | 0.37 |
| **DICModel (ours)** | | **0.72** | **0.19** | **0.58** |
| Intern2.5-VL | 32B | **0.67** | **0.20** | 0.48 |
| Qwen2.5-VL | | 0.65 | **0.20** | 0.46 |

To qualitatively evaluate the output text style of different methods, this section presents the parsing results and the expert- annotated text (as ground-truth) of flowchart. As shown in Fig. 4, the outputs from our DICModel and Gemini-2.0 are similar with the ground-truth text, indicating successful adoption of expert parsing patterns. However, other models with 7B parameters cannot well produce accurate text style. For example, Qwen2.5-VL-7B exhibits logical judgment error and ignores the critical fifth and sixth steps; LLaVA-1.5 7B completely misidentifies the flowchart, directly reproducing the sample in PROMPT. Among the 32B-parameter models, Intern2.5-VL-32B produces text resembling ground truth, while Qwen2.5-VL-32B makes logical error at the fourth step and outputs extraneous asterisks. In summary, our DICModel, the closed-source Gemini-2.0, and Qwen2.5-VL 32B model can successfully mimic the expert's understanding pattern of flowchart.

**Fig. 4.** The parsing visual results of flowchart with different models. Ground-truth is the experts-annotated text according to the input flowchart.

## 5.3 Ablation Study

To validate the effectiveness of the three training stages proposed in this paper, this section conducts ablation studies on visual understanding tasks and visual parsing tasks within the ICT domain. For visual understanding tasks, this section presents the ablation results based on the answer accuracy rate of objective questions constructed by experts; for visual parsing tasks, this paper presents the ablation results based on the BLEU metric. The experimental results are presented in Table 4, where '✓' indicates the inclusion of the training stage, and '×' indicates the exclusion of the training stage.

**Table 4.** Ablation study for the visual question answering and visual parsing tasks. Numbers in bold represent the best scores.

| Training Stages | | | Visual Question Answering | | | Visual Parsing |
|---|---|---|---|---|---|---|
| Stage-1 | Stage-2 | Stage-3 | $Prec_s$ | $Prec_m$ | $Prec_a$ | BLEU |
| × | × | × | 73.0% | 40.0% | 62.0% | 0.37 |
| ✓ | × | × | 79.0% | 39.0% | 65.7% | 0.39 |
| ✓ | ✓ | × | 77.5% | 45.0% | 66.7% | **0.58** |
| ✓ | ✓ | ✓ | **82.5%** | **52.0%** | **72.3%** | 0.13 |

For visual understanding tasks, the three complete training stages proposed in this paper can significantly improve the accuracy rate of DICModel on the evaluation questions constructed by ICT-domain experts. Compared to the base model, DICModel increases the accuracy rate by approximately 9% and 11% on the single-choice and multiple-choice questions, respectively. Therefore, when facing visual understanding tasks, *e.g.*, building a multi-modal vector database, this paper recommends using the DICModel trained after the third stage as the understanding model.

The experimental results indicate that the first and second training stages can help DICModel achieve the highest parsing accuracy with the BLEU score of 0.58. Compared with the baseline model, our DICModel increases the BLEU by approximately 56.8%. This indicates that the first two training stages can significantly enhance the parsing ability of DICModel. Therefore, when solving the image parsing tasks, *e.g.*, delivering semantic image data for LLM pre-training, this paper recommends using the DICModel trained after the second training stage.

## 6 Discussion

This work employs three stages to train a multi-modal LLM to empower the ICT domain. Compared to mixing all training data in a single training phase, the multi-stage training strategy can progressively enhance the image understand capabilities of DICModel. In the first stage, the model focuses on learning fundamental paradigms for converting domain-specific images into structured text. In the second stage, the model learns to emulate domain experts' reasoning patterns to parse images, developing expert-like textual descriptions. In the third stage, the model mainly learns to understand images from global, partial, and detailed dimensions, and can fully explore the high-value information contained within the critical images in the ICT domain.

DICModel can generate detailed text descriptions of images in the ICT domain and answer questions about images, providing a prerequisite for pre-training LLMs and building multi-modal vector databases. Therefore, this work can be applied in the following areas in the future: (1) ICT data asset management, where it generates metadata and descriptions for image databases to facilitate image classification and retrieval; (2) ICT customer service, assisting users in obtaining product details through image-based queries; (3) ICT AI agent development, *e.g.*, multi-modal RAG systems, which can search for related images based on text descriptions or search for related text content based on images, to complete the recommendation of related resources and content. The proposed DICModel achieves robust image understanding performance with only 7B parameters, improving processing efficiency and accuracy while unlocking innovative scenarios for ICT domain. With the development of technology, the application scenarios of DICModel will become more extensive, further promoting the application of artificial intelligence in the ICT industry.

However, this work still has some limitations. The parsing accuracy of DICModel for highly complex flowcharts and signal diagrams remains suboptimal. For flowcharts, when attempting to jump to a target step via a merge line, it often makes mistakes in identifying the target step. For signal diagrams, when a signal steps over multiple nodes to transmit information, it often makes mistakes in identifying the endpoint of the signal. Additionally, while this work optimizes for critical ICT image types, the domain also includes audio and video data, posing challenges for unified cross-modal representation. In the future, we will develop a unified tokenizer for cross-modal integration, enabling a multi-modal model that connects text, audio, images, and video.

## 7 Conclusion

This paper proposes an image understanding and parsing model (DICModel) in the ICT domain. DICModel uses the Qwen2.5-VL 7B model as the base architecture and progressively enhances its captioning ability of the ICT-domain critical images through three data types: synthetic image-text pairs, domain-specific image-text pairs, and visual question answering (VQA) data. Meanwhile, we provide a multi-modal data synthesis pipeline that combines the Mermaid tool and the LLM rewriting component to address training data scarcity. Experimental results indicate that our DICModel is superior to other SOTA models with the same number of parameters, and even surpasses models with several times more parameters, achieving superior performance on both general quantitative metrics and domain expert evaluations. This work may promote

LLM applications in the ICT domain, paving the way for AI agents, digital asset management systems, and intelligent recommendation solutions in this domain.

**Acknowledgments.** xxx

**Disclosure of Interests.** xxx

## References

1. Chen, H., Chen, H., Zhao, Z., et al.: An overview of domain-specific foundation model: key technologies, applications and challenges. arXiv preprint arXiv: 2409.04267 (2024)
2. Wang, B., Xu, C., Zhao, X., et al.: MinerU: An Open-Source Solution for Precise Document Content Extraction. arXiv preprint arXiv:2409.18839 (2024)
3. Wei, H., Liu, C., Chen, J., et al.: General OCR Theory: Towards OCR-2.0 via a Unified End-to-end Model. arXiv preprint arXiv: 2409.01704 (2024)
4. Wei, H., Kong, L., Chen, J., et al.: Vary: Scaling up the vision vocabulary for large vision-language model. In: European Conference on Computer Vision (ECCV), pp. 408-424 (2024)
5. Kulkarni, G., Premraj, V., Ordonez, V., et al.: Babytalk: Understanding and generating simple image descriptions. IEEE transactions on pattern analysis and machine intelligence 35(12), 2891-2903 (2013)
6. Ramos, R., Elliott, D., Martins, B.: Retrieval-augmented image captioning. arXiv preprint arXiv: 2302.08268 (2023)
7. Devlin, J., Chang, M. W., Lee, K., et al.: BERT: Pre-training of Deep Bidirectional Transformers for Language Understanding. In: Proceedings of the 2019 conference of the North American chapter of the association for computational linguistics: human language technologies, pp. 4171-4186 (2019)
8. Vinyals, O., Toshev, A., Bengio, S., et al.: Show and Tell: A Neural Image Caption Generator. In: Proceedings of the IEEE conference on computer vision and pattern recognition, pp. 3156-3164 (2015)
9. Xu, K., Ba, J., Kiros, R., et al.: Show, Attend and Tell: Neural Image Caption Generation with Visual Attention. In: International conference on machine learning. pp. 2048-2057 (2015)
10. Zhou, L., Palangi, H., Zhang, L., et al.: Unified Vision-Language Pre-Training for Image Captioning and VQA. In: Proceedings of the AAAI conference on artificial intelligence, pp.13041-13049 (2020)
11. Li. J., Li, D., Xiong, C., et al.: BLIP: Bootstrapping Language-Image Pre-training for Unified Vision-Language Understanding and Generation. In: International Conference on Machine Learning, pp.12888-12900 (2022)
12. Achiam, J., Adler, S., Agarwal, S., et al.: GPT-4 Technical Report. arXiv preprint arXiv: 2303.08774 (2023)
13. Chen, Z., Wang, W., Cao, Y., et al.: Expanding Performance Boundaries of Open-Source Multimodal Models with Model, Data, and Test-Time Scaling. arXiv preprint arXiv: 2412.05271 (2024)
14. Zhu, J., Wang, W., Chen, Z., et al.: InternVL3: Exploring Advanced Training and Test-Time Recipes for Open-Source Multimodal Models. arXiv preprint arXiv:2504.10479 (2025)
15. Bai, S., Chen, K., Liu, X., et al.: Qwen2.5-VL Technical Report. arXiv preprint arXiv: 2502.13923 (2025)
16. Liu, H., Li, C., Li, Y., et al.: Improved Baselines with Visual Instruction Tuning. In: Proceedings of the IEEE/CVF Conference on Computer Vision and Pattern Recognition, pp. 26296-26306 (2024)
17. Anil, R., Borgeaud, S., Alayrac, J.-B., et al.: Gemini: A Family of Highly Capable Multimodal Models. arXiv preprint arXiv:2312.11805 (2023)
18. Li, J., Li, D., Savarese, S., et al.: BLIP-2: Bootstrapping Language-Image Pre-training with

Frozen Image Encoders and Large Language Models. In: International conference on machine learning, pp.19730-19742 (2023)
19. Yang, X., Tang, K., Zhang, H., et al.: Auto-Encoding Scene Graphs for Image Captioning. In: Proceedings of the IEEE/CVF conference on computer vision and pattern recognition, pp.10685-10694 (2019)
20. Zhong, Y., Wang, L., Chen, J., et al. Comprehensive Image Captioning via Scene Graph Decomposition. In: 16th European Conference, pp. 211-229 (2020)
21. Yao, T., Pan, Y., Li, Y., et al.: Exploring Visual Relationship for Image Captioning. In: Proceedings of the European conference on computer vision, pp. 684-699 (2018)
22. Herdade, S., Kappeler, A., Boakye, K., et al.: Image Captioning: Transforming Objects into Words. In: Advances in neural information processing systems, arXiv preprint arXiv:1906.05963 (2019)
23. Huang, L., Wang, W., Chen, J., et al.: Attention on attention for image captioning. In: Proceedings of the IEEE/CVF international conference on computer vision, pp. 4634-4643 (2019)
24. Mokady, R., Hertz, A., Bermano, A., H.: ClipCap: CLIP Prefix for Image Captioning. arXiv preprint arXiv: 2111.09734 (2021)
25. Qi, S., Cao, Z., Rao, J., et al.: Understanding Multimodal LLMs: the Mechanistic Interpretability of LLaVA in Visual Question Answering. arXiv preprint arXiv:2411.10950 (2024)
26. Radford, A., Kim, J. W., Hallacy, C., et al.: Learning Transferable Visual Models From Natural Language Supervision. In: International conference on machine learning. pp. 8748-8763 (2021)
27. Alayrac, J. B., Donahue, J., Luc, P., et al.: Flamingo: a Visual Language Model for Few-Shot Learning. In: Advances in neural information processing systems, pp. 23716-23736 (2022)
28. Contributors, O., C.: OpenCompass: A universal evaluation platform for foundation models. URL: https://github. com/open-compass (2023)
29. Sun, X., Chen, Y., Huang, Y., et al.: Hunyuan-large: An open-source moe model with 52 billion activated parameters by tencent. arXiv preprint arXiv: 2411.02265 (2024)
30. Zeng, A., Xu, B., Wang, B. et al.: ChatGLM: A Family of Large Language Models from GLM-130B to GLM-4 All Tools. arXiv preprint arXiv:2406.12793 (2024)
31. Seed, B., Chen, J., Fan, T., et al.: Seed1.5-Thinking: Advancing Superb Reasoning Models with Reinforcement Learning. arXiv preprint arXiv: 2504.13914 (2025)
32. Grok, X. Beta.: Grok 3 Beta-The Age of Reasoning Agents. URL: https://x. ai/news/grok-3 (2025)
33. Enis, M., Hopkins, M.: From LLM to NMT: Advancing Low-Resource Machine Translation with Claude. arXiv preprint arXiv:2404.13813 (2024)
34. Abdin, M., Aneja, J., Awadalla, H., et al.: Phi-3 Technical Report: A Highly Capable Language Model Locally on Your Phone Visual Instruction Tuning. arXiv preprint arXiv: 2404.14219 (2024)
35. Liu, H., Li, C., Li, Y., et al.: Improved baselines with visual instruction tuning. In: Proceedings of the IEEE/CVF Conference on Computer Vision and Pattern Recognition, pp. 26296-26306 (2024)
36. Chen, Z., Wang, W., Cao, Y., et al.: Expanding performance boundaries of open-source multimodal models with model, data, and test-time scaling. arXiv preprint arXiv:2412.05271 (2024)